\documentclass[10pt,twocolumn,letterpaper]{article}

\usepackage{cvpr}              

\newcommand{\OURNAME}{AesthetiQ}

\definecolor{cvprblue}{rgb}{0.21,0.49,0.74}
\usepackage{amsmath}
\usepackage{graphicx} 
\usepackage[pagebackref,breaklinks,colorlinks,allcolors=cvprblue]{hyperref}
\usepackage{multirow}

\def\paperID{10036} 
\def\confName{CVPR}
\def\confYear{2025}

\title{\textit{\OURNAME}: Enhancing Graphic Layout Design via Aesthetic-Aware Preference Alignment of Multi-modal Large Language Models}

\author{Sohan Patnaik\footnote{rishabhj@adobe.com}
\\
MDSR Adobe\\
\and
Rishabh Jain\footnote{rishabhj@adobe.com}\\
MDSR Adobe\\
\and
Balaji K\\
Adobe Research\\
\and
Mausoom Sarkar\\
MDSR Adobe\\
}

\author{Sohan Patnaik\thanks{equal contribution}\\
MDSR Adobe\\
\and
Rishabh Jain\footnotemark[1]\\
MDSR Adobe\\
\and
Balaji Krishnamurthy\\
MDSR Adobe\\
\and
Mausoom Sarkar\\
MDSR Adobe\\
}

\begin{document}
\maketitle

\begin{abstract}
    Visual layouts are essential in graphic design fields such as advertising, posters, and web interfaces. The application of generative models for content-aware layout generation has recently gained traction. However, these models fail to understand the contextual aesthetic requirements of layout design and do not align with human-like preferences, primarily treating it as a prediction task without considering the final rendered output. To overcome these problems, we offer \textbf{Aesthetic-Aware Preference Alignment} (AAPA), a novel technique to train a Multi-modal Large Language Model (MLLM) for layout prediction that uses MLLM's aesthetic preferences for Direct Preference Optimization over graphic layouts. We propose a data filtering protocol utilizing our layout-quality heuristics for AAPA to ensure training happens on high-quality layouts. Additionally, we introduce a novel evaluation metric that uses another MLLM to compute the win rate of the generated layout against the ground-truth layout based on aesthetics criteria. We also demonstrate the applicability of AAPA for MLLMs of varying scales (1B to 8B parameters) and LLM families (Qwen, Phi, InternLM). By conducting thorough qualitative and quantitative analyses, we verify the efficacy of our approach on two challenging benchmarks - Crello and Webui, showcasing $17\%$, and $16\%$ improvement over current State-of-The-Art methods, thereby highlighting the potential of MLLMs in aesthetic-aware layout generation.
\end{abstract}    
\section{Introduction}
\label{sec:intro}

\begin{figure}[t!]
\begin{center}
  \includegraphics[width=\linewidth]{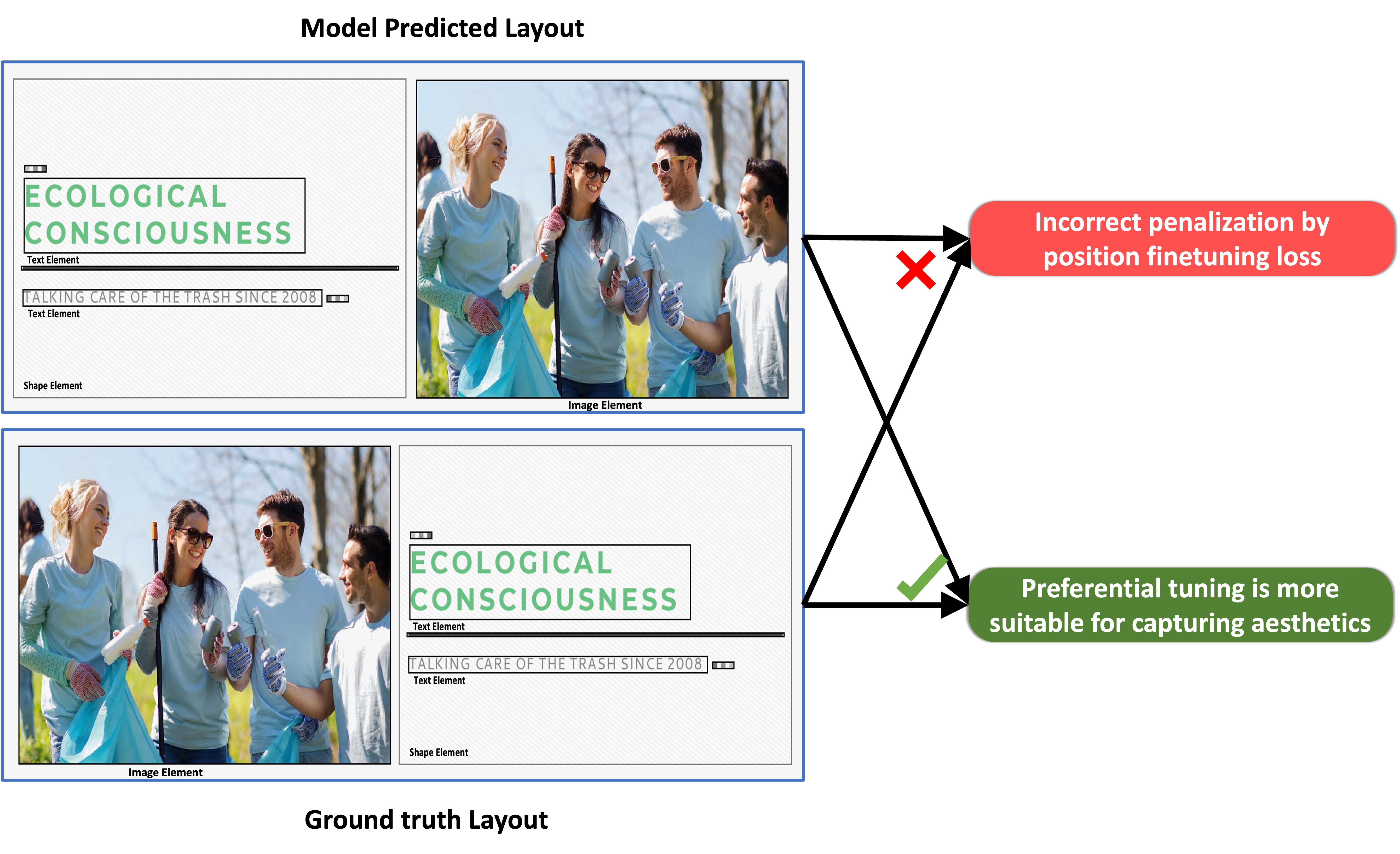}
\end{center}
\vspace{-5mm}
 \caption{Existing cross-entropy loss based methods penalize element misalignment heavily while preferential tuning via AAPA better capture aesthetic nuances in layouts} 
\label{fig:hero_fig}
\end{figure}
The arrangement of visual elements for a graphic layout significantly impacts how users perceive and interact with digital content~\citep{10.1007/978-3-030-58580-8_29}. The growing demand for automated design solutions demonstrates broad impact across multiple domains, including poster design~\citep{Yamaguchi2021CanvasVAELT, inoue2023document}, user interface design~\citep{10.1145/3126594.3126651, Jiang_Sun_Zhu_Lou_Zhang_2022}, document layouts~\citep{10.1145/3306346.3322971, Yamaguchi2021CanvasVAELT}, and presentation slides~\citep{Fu2021DOC2PPTAP}. Designers expect layouts to achieve dual-level harmony, creating a seamless blend of fine-grained design elements - color, font, and layout - while ensuring they work cohesively with different types of elements in the overall design~\citep{huang2023surveygraphicdesignintelligence}. These fine-grained components must maintain aesthetic integrity and enhance the design's collective harmony. For instance, when examining basic design elements, colors must strike a delicate balance between contrast and cooperation, delivering visual style.

As digital content creation accelerates, the field of automated layout generation has evolved across multiple directions. Research has progressed from unconditional generation~\citep{Arroyo_2021_CVPR, Gupta2020LayoutTransformerLG, Jiang_Sun_Zhu_Lou_Zhang_2022, li2018layoutgan, Yamaguchi2021CanvasVAELT} to more sophisticated approaches, including conditional generation based on user inputs (such as element types~\citep{10.1145/3474085.3475497, 10.1007/978-3-031-19790-1_29, 10.1007/978-3-030-58580-8_29}, sizes~\citep{10.1007/978-3-031-19790-1_29, li2018layoutgan}, or relationships~\citep{10.1145/3474085.3475497, 10.1007/978-3-030-58580-8_29}), conditional refinement using coarse attributes \citep{10.1145/3397482.3450716}, and conditional completion with partially available elements~\citep{Gupta2020LayoutTransformerLG}. Recently, Multi-modal Large Language Models (MLLMs) have demonstrated remarkable capabilities in spatial reasoning and visual-semantic understanding~\citep{Chen2023InternVS, Liu2023VisualIT, openai2024gpt4technicalreport}. Their ability to process both textual and visual information enables them to capture complex spatial relationships and translate challenging language concepts into coherent layout arrangements. Current approaches~\citep{10.1007/978-3-031-73007-8_26, yang2024posterllava, tang2024layoutnuwa} primarily reduce the complex design problem to a numeric representation, where each design element is simplified to a tuple $(x,y,w,h)$ encoding its center coordinates, width, and height. While both autoregressive models and diffusion models~\cite{Zhang2023LayoutDiffusionIG, Arroyo2021VariationalTN, Chai_2023_CVPR} approach layout generation differently-the former through sequential prediction and the latter through denoising-they share a fundamental limitation: both treat layouting as an optimization problem over position representations. As illustrated in Fig.\ref{fig:hero_fig}, when multiple layouts are aesthetically pleasing, optimizing based on the absolute numerical position through cross-entropy-based loss incorrectly penalizes other alternative layouts predicted by the model, thus enforcing incorrect understanding. Moreover, reducing visual elements to numerical values strips away critical design attributes, hierarchical relationships, and semantic context that human designers naturally consider. This reductionist approach fails to capture the aesthetics or \textit{visual appeal} of the rendered layouts, particularly when dealing with complex layouts that require subtle aesthetic considerations and contextual awareness.

To this end, we pose a fundamental research question: \textit{``Can we utilize aesthetic preferences to guide the training of MLLMs, thereby enhancing the visual impact of generated layouts?"} While MLLMs demonstrate strong visual-semantic alignment capabilities, these are often underutilized when trained solely with supervised position loss per element, which enforces a single layout variation for each input. Although incorporating aesthetic awareness through rendered layouts and an aesthetic discriminator seems intuitive, creating a differentiable layout renderer and the discriminator is non-intuitive. Instead, we propose learning aesthetics by proxy through preference learning between multiple layouts, hypothesizing that this approach enables MLLMs to develop a deeper contextual and aesthetic understanding of design principles.

In this paper, we introduce \textbf{\textit{Aesthetic-Aware Preference Alignment}} (AAPA), a novel algorithm that extends Direct Preference Optimization (DPO)~\citep{rafailov2023direct} for aesthetic-aware layout generation. AAPA creates multiple layout candidates and leverages an MLLM as a judge to rank them based on their aesthetics, thereby developing robust aesthetic preferences. To ensure reliable training, we implement a rigorous data filtering protocol using alignment and overlap metrics, guaranteeing the model learns preferences only from well-formed layouts. Additionally, we present a novel evaluation framework using MLLMs as aesthetic scorers, introducing the \textit{\% win rate} metric that measures how often model-generated layouts outperform ground-truth layouts aesthetically. This approach offers more direct quality assessment than traditional metrics like Intersection over Union (IoU) or Boundary Displacement Error (BDE)~\citep{9525300}, which fail to capture subtle aesthetic considerations. Our main contributions can be summarized as follows:
\begin{itemize}
    \item We demonstrate that VLMs can effectively utilize their real-world knowledge and visual-semantic alignment capabilities to enhance layout prediction tasks.
    \item We propose Aesthetic-Aware Preference Alignment, a novel method to improve aesthetics of generated layouts through preference-based learning.
    
    \item We highlight the importance of data filtering based on quality heuristics through alignment and overlap heuristics, ensuring robust training on high-quality samples.
    \item We develop a novel evaluation framework that assess layout aesthetics via MLLMs, providing more meaningful quality measurements than traditional metrics.
    \item We conduct extensive qualitative and quantitative analyses of each proposed component using two challenging benchmark datasets, Crello and WebUI, demonstrating the effectiveness of our approach across diverse settings.
    \item We showcase the versatility and effectiveness of our approach across MLLMs of varying scales and families, and perform several ablation studies to provide deeper insights into the contributions of each component.
\end{itemize}

\section{Related works}
\label{sec:related_work}

\textbf{Graphic Layout Generation:} As digital content creation continues to grow rapidly, the automated generation of visually appealing layouts that meet users' needs has emerged as an important research problem. Early works employed rule-based~\citep{6777138} and energy optimization~\citep{10.1145/2702123.2702149, 10.1007/978-3-030-58580-8_29} techniques to structure graphic layouts, providing a solid foundation but offering limited flexibility. GAN-based approaches~\citep{10.1109/TVCG.2020.2999335,10.1145/3474085.3475497, Zhou2022CompositionawareGL} introduced frameworks for unconditional layout generation by capturing geometric relationships across elements and leveraging differentiable rendering in the image-space, while transformer models such as LayoutTransformer~\citep{Gupta2020LayoutTransformerLG}, VTN~\citep{Arroyo2021VariationalTN}, and BLT~\citep{10.1007/978-3-031-19790-1_29} used autoregressive and bidirectional transformers to improve layout quality. With advancements in NLP, masking strategies, language models, and encoder-decoder architectures have been applied to layout generation. Approaches like CanvasVAE~\citep{Yamaguchi2021CanvasVAELT}, FlexDM~\citep{inoue2023document}, and ICVT~\citep{10.1145/3503161.3548332} integrate images and text to enhance content relevance, enabling conditional generation through multi-modal features extracted via transformer-based architectures~\citep{10.5555/3295222.3295349, Devlin2019BERTPO, lewis-etal-2020-bart}. However, these methods still face challenges in adaptability to diverse content and impose constraints that limit their practical usability in real-world design applications.

\vspace{0.5mm}
\noindent
\textbf{Constraint-aware Layout Prediction:} Recent works have leveraged diffusion models to improve layout generation quality across various conditions. PLay~\citep{10.5555/3618408.3618624} uses a latent diffusion model conditioned on guidelines to align elements, while LayoutDM~\citep{Chai_2023_CVPR} and LDGM~\citep{Hui2023UnifyingLG} employ the discrete diffusion framework~\citep{austin2021structured} to build unified models without guidelines, using attribute-specific corruption strategies~\citep{9879180} to restrict variables of different attributes to their respective sample spaces. LDGM further applies discretized Gaussian noise to enable gradual coordinate changes. Following a similar design, LayoutDiffusion~\citep{Zhang2023LayoutDiffusionIG} and LACE~\citep{chen2024towards} enhance visual quality by scaling up the transformer backbone and improving alignment and overlap metrics. However, these methods lack contextual understanding and fail to capture semantic relationships across elements, leading to designs that may be geometrically accurate but aesthetically incoherent. Other diffusion-based methods~\citep{10.1007/978-3-031-41676-7_21, 10204668} attempt to generate layouts in continuous space but still rely on quantized geometric attributes or conditional generation based on categorical attributes, limiting their flexibility. Training diffusion models with masked and unmasked layouts~\citep{Wei2023DiffusionMA} has shown potential for unifying various tasks but still lacks the contextual understanding.

\vspace{0.5mm}
\noindent
\textbf{LLMs Assisted Layout Generation:} Recent works have explored large language models (LLMs) for layout generation, treating layouts as structured data formats like XML or JSON. LayoutNUWA~\citep{tang2024layoutnuwa} fine-tunes LLaMa~\citep{Touvron2023LLaMAOA} and CodeLLaMa~\citep{Rozire2023CodeLO} for content-agnostic layout generation, achieving state-of-the-art results across multiple datasets. LayoutPrompter~\citep{10.5555/3666122.3668024} introduces a training-free approach by leveraging RAG (Retrieval-Augmented Generation) to enhance in-context learning in GPT~\citep{10.5555/3495724.3495883}, though it is limited to open-domain generation. These methods, however, translate visual domain features into hard tokens, potentially leading to significant information loss. To address this, recent works propose multi-modal techniques such as visual instruction tuning~\citep{liu2023visual} with aligned visual adaptation heads, allowing models to process visual information directly. While these approaches show promise, they rely on cross-entropy loss for optimization, which penalizes aesthetically good layouts along with poor ones (as depicted in Fig.\ref{fig:hero_fig}), neglecting aesthetic preferences and leading to suboptimal design outcomes. This cross-entropy loss-based optimization fails to capture nuanced aesthetic considerations essential for high-quality layouts.

\section{Methodology}

\begin{figure*}[h]
    \centering
  \includegraphics[width=0.9\linewidth]{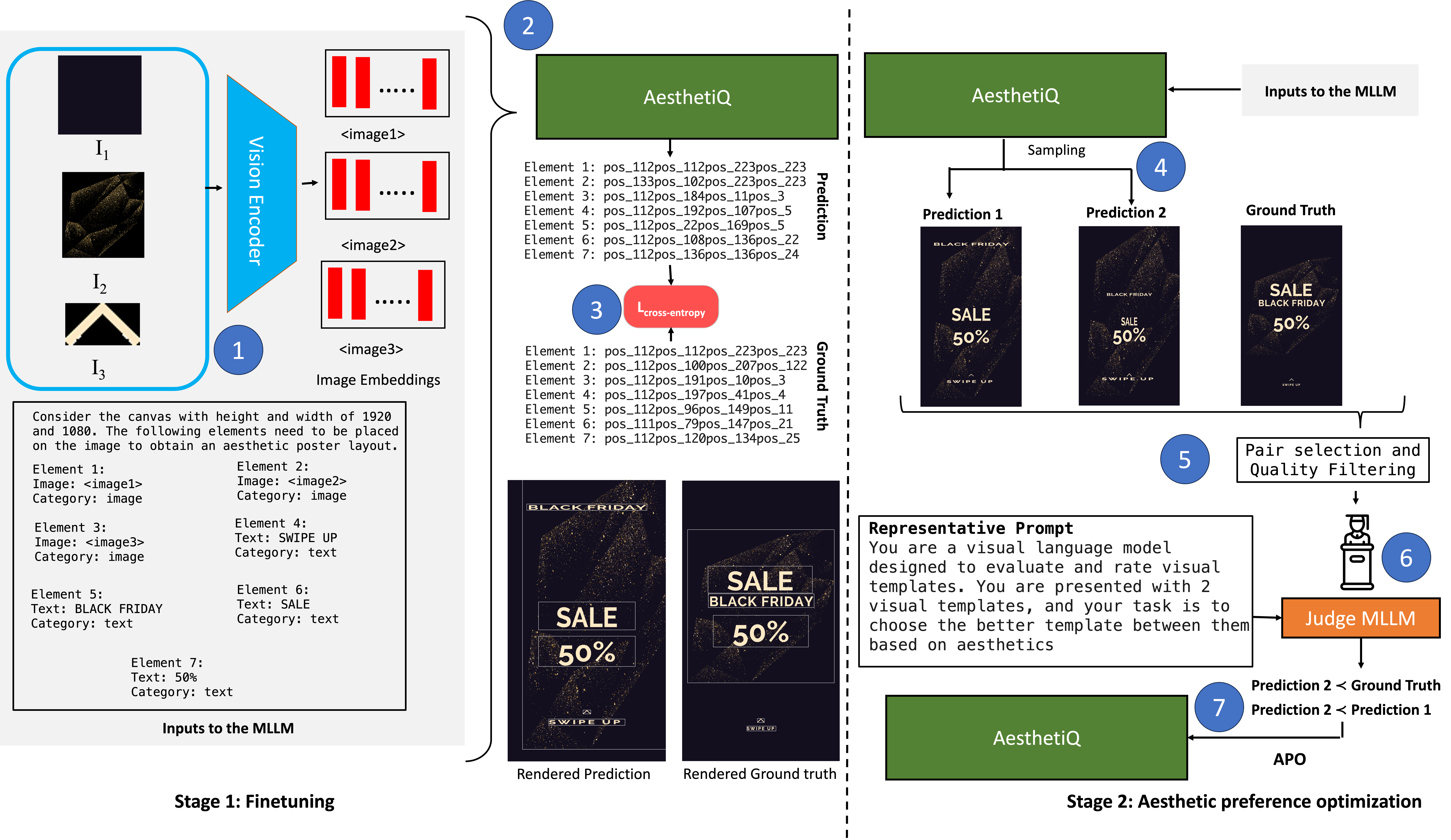}
  \caption{The training for the aesthetic layout prediction task consists of the following steps: 
1) \textbf{Vision Encoder}: Design elements (images and text) are processed to generate image and text embeddings. 
2) \textbf{\OURNAME\ Model Prediction}: Embeddings are passed to the \OURNAME\ model, which predicts layout coordinates. 
3) \textbf{Training with Cross-Entropy Loss}: The predicted layout is compared with the ground truth and trained using cross-entropy loss. 
4) \textbf{Sampling for Comparison}: Multiple layout predictions are generated using \OURNAME\ inference. 
5) \textbf{Pair Selection and Quality Filtering}: We filter the data based on quality heuristics to ensure layout quality in samples. 
6) \textbf{Judging by ViLA}: The ViLA model compares layout pairs and selects the better one based on aesthetic preferences. 
7) \textbf{Aesthetic Preference Optimization (AAPA)}: Feedback from ViLA is used to fine-tune the \OURNAME\ model for aesthetic optimization.}
  \label{architectureDiagram}
\end{figure*}

Our approach to layout prediction focuses on generating aesthetically aligned designs, with a Multi-modal Large Language Model (MLLM) serving as an evaluator for layout quality. We want to capture design aesthetics in layout configurations by using another MLLM (judge). To address the challenges given by subjective design preferences, limited datasets, and high-dimensional variability, we use an Aesthetic-Aware Preference Alignment (AAPA) technique, which aligns our model's output with aesthetic evaluations from the judge MLLM.  We begin by formalizing the layout prediction task, discussing the input and output representations, and the fine-tuning approach. We then introduce AAPA, which drives aesthetic sensitivity in layout generation (Fig \ref{architectureDiagram}). Finally, we present additional insights gained from scaling and pretraining on a large dataset of templates.

\subsection{Layout Prediction Task}
The goal of layout prediction involves placing design elements over a blank canvas to generate an aesthetically pleasing graphic template. Formally, consider a set of $N$ design elements $E = \{e_1, e_2, \cdots, e_N\}$ that need to be placed over a canvas of width $W_c$ and height $H_c$ to obtain a graphic layout $G$. We define the placement of an element $e_i$ through bounding box attributes $b_i = (x_i, y_i, w_i, h_i)$ that represent the graphic element's center coordinates, width, and height of the bounding box. Given the type of each element denoted by $T(e_i) \in \mathcal{T}$, where $\mathcal{T} = \{\text{image}, \text{text}, \text{shape}, \text{background}\}$, we verbalise the input through a prompt template denoted by $\mathcal{P}^E_{H_c, W_c}$ (Eq.\ref{eqn:prompt}).
\begin{equation}
    \mathcal{P}^E_{H_c, W_c} = \bigcup_{i=1}^{n} \begin{cases} 
    \text{ImageTokens}(e_i) & \text{if } T(e_i) \in \mathcal{T} - \{\text{text}\} \\ 
    \text{TextTokens}(e_i) & \text{if } T(e_i) = \text{text} 
\end{cases}
\label{eqn:prompt}
\end{equation}

Consider the layout MLLM \(\mathcal{M}_{\text{layout}}\) with three components: a vision encoder \( f_{\text{vision}} \), a text encoder \( f_{\text{text}} \), and a large language model \( \mathcal{L} \). $\mathcal{M}_{\text{layout}}$ takes as input the verbalized prompt, \(\mathcal{P}^E_{H_c, W_c}\), by encoding each token in the prompt using either a vision encoder or a text encoder, depending on the type of the corresponding design element. For each token representing an element \(e_i\) where \( T(e_i) \in \mathcal{T} - \{\text{text}\} \), the vision encoder \( f_{\text{vision}} \) produces an embedding \( z_i^{\text{vision}} = f_{\text{vision}}(\text{ImageTokens}(e_i)) \). For tokens representing elements where \( T(e_i) = \text{text} \), the text encoder \( f_{\text{text}} \) produces an embedding \( z_i^{\text{text}} = f_{\text{text}}(\text{TextTokens}(e_i)) \). The final multi-modal embedding \( Z \) for \(\mathcal{P}^E_{H_c, W_c}\) is then represented as the union of all encoded tokens, combining vision and text embeddings based on each token’s type (Eq.\ref{eqn:embeddings}).
\begin{equation}
    Z = \bigcup_{i=1}^{N} \begin{cases} 
       z_i^{\text{vision}} & \text{if } T(e_i) \in \mathcal{T} - \{\text{text}\} \\ 
       z_i^{\text{text}} & \text{if } T(e_i) = \text{text} 
   \end{cases}
\label{eqn:embeddings}
\end{equation}

Once the multi-modal embedding \( Z \) is constructed, it is fed into the large language model component \( \mathcal{L} \) of \(\mathcal{M}_{\text{layout}}\) to predict the position of each design element. 

\subsection{Position-Aware Layout Instruction Tuning}
To facilitate accurate layout prediction, we discretize the bounding box coordinates \( b_i = (x_i, y_i, w_i, h_i) \). These continuous attributes are mapped into discrete bins based on the width \( W_c \) and height \( H_c \) of the canvas. The binning function \( B(\cdot) \) depicted by Eq. \ref{eqn:bin_func} maps each continuous attribute into one of \( K \) bins, where \( K \) is the number of bins:
\vspace{-0.5mm}
\begin{equation}
B(a_i) = \left\lfloor \frac{a_i}{D} \times K \right\rfloor \quad \text{with} \quad D = \begin{cases} 
W_c & \text{if } a_i \in \{x_i, w_i\} \\
H_c & \text{if } a_i \in \{y_i, h_i\}
\end{cases}
\label{eqn:bin_func}
\end{equation}
For each of these binned values, we introduce position tokens \( \text{pos}_k \), where \( k \in \{0, 1, \dots, K\} \). Thus, each attribute \( x_i, y_i, w_i, h_i \) is assigned the tokens $\text{pos}_{B(x_i)}$, $\text{pos}_{B(y_i)}$, $\text{pos}_{B(w_i)}$, and $\text{pos}_{B(h_i)}$ respectively. This discretization reduces the continuous spatial attributes to a fixed set of discrete position tokens, simplifying the layout prediction task. By quantizing the position, the model can more effectively learn the spatial relationships across design elements. Once the position tokens are integrated into the multi-modal embedding, the language model component \( \mathcal{L} \) predicts the layout by outputting the corresponding position tokens for each element, which are mapped back to approximate spatial coordinates using the function $B^{-1}$.

Let \( Y = \{y_1, y_2, \dots, y_M\} \) represent the ground truth sequence of position tokens corresponding to the predicted bounding box attributes of the design elements, where each \( y_i \in \{ \text{pos}_0, \text{pos}_1, \dots, \text{pos}_K \} \), and \( M \) is the number of output tokens. $\mathcal{L}$ outputs policy \(\pi_{\mathcal{M}} (\mathcal{P}_{H_c,W_c}^{E}) \), a sequence of logits \( \hat{y}_i \), where \( \hat{y}_i \in \mathbb{R}^K \) represents the predicted probabilities for each position token of each design element \( e_i \). We train $\mathcal{M}_{\text{layout}}$ through the cross-entropy loss (Eq. \ref{eqn:ce_loss})
\vspace{-0.5mm}
\begin{equation}
\mathcal{L}_{\text{CE}} = - \frac{1}{S} \sum_{i=1}^{S}\sum_{j=1}^{M} \sum_{k=0}^{K} \mathbb{I}(y_j = \text{pos}_k) \log \hat{y}_{j,k}
\label{eqn:ce_loss}
\end{equation}
where \( \mathbb{I}(y_i = \text{pos}_k) \) is the indicator function and \( \hat{y}_{i,k} \) is the predicted probability (logit) for the \( k \)-th position token of the \( j \)-th element, and S is the size the dataset. Thus, during instruction tuning, the $\mathcal{M}_{\text{layout}}$ learns the correct spatial placement of each element through the alignment of the predicted position tokens with the ground truth tokens. However, cross-entropy loss is suboptimal as it requires an exact match (Fig \ref{fig:hero_fig}). Additionally, if the ground-truth position token is $pos_{32}$ and the model predicts $pos_{33}$, the cross-entropy loss penalizes this heavily despite the minimal difference between the two positions.

\subsection{Aesthetic-Aware Preference Alignment}
A significant challenge in layout prediction arises from the non-differentiable nature of rendering individual elements on a canvas, which prevents the model from directly ``\textit{seeing}" the rendered template for aesthetic evaluation.  To address this, we introduce a judge model \( \mathcal{M}_{\text{judge}} \) that performs a pairwise comparison of two graphic layouts to determine which one is aesthetically superior. Given two graphic layouts, rendered on a canvas using a renderer $\mathcal{R}$, denoted as \( G^i = \mathcal{R}(E^i, {W_c}^i, {H_c}^i)\), \( i = \{1, 2\} \), the goal of \( \mathcal{M}_{\text{judge}} \) is to decide which layout is better based on aesthetic criteria. $\mathcal{M}_{\text{judge}}$ takes as input a judge prompt \( \mathcal{P}_{\text{judge}} \), which verbalizes the evaluation criteria for comparing the two layouts, and the two graphic layouts \( G^1 \) and \( G^2 \), to produce a binary decision about which layout is better. Formally, 
\begin{equation}
d = \mathcal{M}_{\text{judge}}(\mathcal{P}_{\text{judge}}, G^1, G^2)
\end{equation}
where \( d \in \{1, 2\} \) represents the decision of the $\mathcal{M}_{judge}$. If \( d = 1 \), then graphic layout \( G^1 \) is considered better than graphic layout \( G^2 \), and vice versa. Thus, we define
\begin{equation}
    G^{\text{w}} = 
    \begin{cases} 
    G^1 & \text{if } d = 1 \\ 
    G^2 & \text{if } d = 2 
    \end{cases}, 
\quad
G^{\text{r}} = 
    \begin{cases} 
    G^2 & \text{if } d = 1 \\ 
    G^1 & \text{if } d = 2 
    \end{cases}
\end{equation}

Through this pairwise comparison, the judging model \( \mathcal{M}_{\text{judge}} \) selects the best ($G^w$) and worst ($G^r$) graphic layouts based on aesthetic quality without computing any explicit scores. Scoring individual layouts was found to be challenging, as MLLMs often assign similar scores across a range of layouts due to their limited exposure to the design space. Once we obtain the preferential data from \( \mathcal{M}_{\text{judge}} \), $\mathcal{M}_{\text{layout}}$ is tuned further to prefer the better layout compared to the worse layout evaluated based on aesthetics. For this, we apply Aesthetic-Aware Preference Alignment (AAPA) to train the layout model \( \mathcal{M}_{\text{layout}} \) using AAPA loss (Eq. \ref{eqn:dpo_loss})
\vspace{-0.9mm}
\begin{equation}
\resizebox{\columnwidth}{!}{$
    \mathcal{L}_{\text{AAPA}} = -\log \sigma \Bigg (\beta \Big( \log \frac{\pi_{\mathcal{M}} (\mathcal{P}_{H_c,W_c}^{E^{w}})}{\pi_{\hat{\mathcal{M}}} (\mathcal{P}_{H_c,W_c}^{E^{w}})} - \log \frac{\pi_{\mathcal{M}} (\mathcal{P}_{H_c,W_c}^{E^{r}})}{\pi_{\hat{\mathcal{M}}} (\mathcal{P}_{H_c,W_c}^{E^{r}})} \Big ) \Bigg)
$}
\label{eqn:dpo_loss}
\end{equation}
where $\pi_{\mathcal{M}}$ and $\pi_{\mathcal{\hat{M}}}$ are the policies of layout prediction and reference model respectively, and $\beta=0.1$ is a hyper-parameter to control divergence from the reference model. We initialize the reference model $\mathcal{\hat{M}}$ to be the same as the layout prediction model $\mathcal{M}_{\text{layout}}$ at the start of the training.

\vspace{1.5mm}
\noindent
\textbf{Quality Metrics-based Layout Filtering}: To ensure high quality layouts are used to train $\mathcal{M}_{\text{layout}}$ through AAPA, we enforce a data filtration process based on quality metrics that evaluate the alignment and overlap of the predicted bounding boxes. The goal is to maximize alignment between design elements while minimizing overlap, ensuring that the selected layouts are aesthetically pleasing and well-structured. For each design element \( e_i \), let \(\theta_i = (x_i^L, y_i^T, x_i^C, y_i^C, x_i^R, y_i^B)\) represents the top-left (\(x_i^L, y_i^T\)), center (\(x_i^C, y_i^C\)), and bottom-right (\(x_i^R, y_i^B\)) coordinates of the bounding box. We compute the minimum distance between the key coordinates of element $e_i$ and all of those other elements \( e_j \) (\( j \neq i \)), to find the adjacent element with respect to the key coordinates and compute alignment quality metric $\mathcal{Q}_{\text{align}}$, depicted by Eq.\ref{eqn:alignment}.
\begin{equation}
    \mathcal{Q}_{\text{align}} (G) = \frac{1}{N} \sum_{i=1}^{N} \frac{\min (f(\Delta x_i^*), f(\Delta y_i^*)) - 1}{e - 1} 
    \label{eqn:alignment}
\end{equation}
where \( f(x) = \exp(1 - x) \), \( \Delta x_i^* \in \{\Delta x_i^L, \Delta x_i^C, \Delta x_i^R\} \) and \( \Delta y_i^* \in \{\Delta y_i^T, \Delta y_i^C, \Delta y_i^B\} \). The horizontal and vertical distances are computed as $\Delta x^*_i = \min_{j \neq i} |x^*_i - x^*_j|$, and $\Delta y^*_i = \min_{j \neq i} |y^*_i - y^*_j|$ respectively. Similarly, we compute the average overlap of an element $e_i$ with all other elements $e_j$ and define $\mathcal{Q}_{\text{overlap}}$ using Eq. \ref{eqn:overlap}. 
\begin{equation}
    \mathcal{Q}_{\text{overlap}}(G) = \frac{1}{N} \sum_{i=1}^{N} \sum_{j \neq i} \Big (1 - \frac{\text{Area}(e_i \cap e_j)}{\text{Area}(e_i)} \Big )
    \label{eqn:overlap}
\end{equation}
where, $\text{Area}(e_i \cap e_j)$ denotes the area of intersection between element $e_i$ and $e_j$ and $\text{Area}(e_i)$ denotes the area of element $e_i$. Ideally, we want overlap to be less, and therefore higher $\mathcal{Q}_{\text{overlap}}$ denotes lesser overlap. We compute the overall quality metric \( \mathcal{Q}(G) \) for a graphic layout \( G \) by taking the average of the alignment score \( \mathcal{Q}_{\text{align}} \) and the overlap score \( \mathcal{Q}_{\text{overlap}} \). For a given layout \( G_s \), if its quality metric satisfies \( \mathcal{Q}(G_s) > \mu_{\mathcal{Q}} - \sigma_{\mathcal{Q}} \), i.e., the quality metric exceeds one standard deviation below the mean quality metric across the entire dataset, we classify it as a higher-quality layout. By applying this data filtration process, we enhance the overall quality of the training data, which in turn helps to effectively train the layout model \( \mathcal{M}_{\text{layout}} \).

\vspace{1mm}
\noindent
\textbf{Aesthetic-Aware Layout Evaluation:} To evaluate the aesthetic quality of predicted layouts, we employ the judge model $\mathcal{M}_{\text{judge}}$ to perform pairwise comparisons between predicted layouts and their corresponding ground truth layouts. For a test dataset $\mathcal{D}_{\text{test}}$ containing $S$ samples, let $G_p^i = \mathcal{R}(\pi_{\mathcal{M}}(\mathcal{P}_{H_c,W_c}^{E^i}), W_c^i, H_c^i)$ be the predicted layout and $G_g^i = \mathcal{R}(E^i, W_c^i, H_c^i)$ be the ground truth layout for the $i$-th sample. The win rate $\mathcal{W}$ is computed as:
\begin{equation}
    \mathcal{W} = \frac{1}{S} \sum_{i=1}^S \mathbb{I} \Big\{ \mathcal{M}_{\text{judge}} (\mathcal{P}_{\text{judge}}, G_p^i, G_g^i) = 1\Big\}
\end{equation}
where $\mathbb{I}(\cdot)$ is the indicator function that returns 1 if the predicted layout wins over the ground truth layout in the aesthetic comparison, and 0 otherwise. A higher win rate indicates that the layout model $\mathcal{M}_{\text{layout}}$ generates layouts that are aesthetically superior to the ground truth layouts.

\subsection{Pre-training to Enhance Layout Understanding}
To further explore the role of data size for pretraining data hungry MLLM's, we trained our model on an 80,000 template dataset collected from an online design creation platform. This pretraining phase aims to imbue the model with foundational design principles, such as minimizing element overlap, enhancing alignment, and prioritizing salient regions for important elements. Our experiments reveal that pretraining before position-aware layout instruction tuning significantly boosts the model's aesthetic sensitivity, as evidenced by improvements in alignment with design principles when compared to models trained without pretraining.

\section{Experimental Details}

\textbf{Datasets:} We evaluate \OURNAME\ on two benchmark graphic layout datasets: Crello~\citep{Yamaguchi2021CanvasVAELT} and WebUI~\citep{10655522}, both posing unique challenges. Crello dataset contains 23,302 vector-based design templates across various formats (e.g., social media posts, banners), split into 19,479 training, 1,852 validation, and 1,971 test samples. Its primary challenge lies in handling diverse canvas sizes and aspect ratios while maintaining design consistency. WebUI dataset comprises 70k web page UIs, including visual screenshots and metadata. WebUI's challenges stem from variability in web page structures across domains. Both datasets test the model's ability to manage complex layouts and ensure semantic coherence across diverse design structures.


\vspace{0.75mm}
\noindent
\textbf{Implementation Details:} For our layout prediction model $\mathcal{M}_{\text{layout}}$, we use InternVL~\citep{Chen2023InternVS} as the backbone, given its robust spatial understanding and visual reasoning capabilities across diverse tasks like DocVQA~\citep{9423358}, TextVQA~\citep{8953586}, MME~\citep{Fu2023MMEAC}, MMMU~\citep{Yue2023MMMUAM}, and MMVET~\citep{yu2024mm}. InternVL’s multi-image training aligns well with layout prediction, where interpreting contextual relationships between multiple elements is crucial. For judging layouts based on aesthetics, we leverage VILA-7B ~\citep{10657989} as $\mathcal{M}_{\text{judge}}$. We train \OURNAME\ for 20 epochs on an effective batch size of 128 using 8 80GB A100 GPUs. Position coordinates are discretized with $K=224$ tokens, and $\mathcal{M}_{\text{layout}}$ is optimized using a learning rate of $4e^{-5}$, weight decay of $0.01$, and Cosine Annealing~\citep{loshchilov2017sgdr} with warmup ratio $0.03$ through the AdamW optimizer~\citep{loshchilov2018decoupled}. Layout preferences for AAPA are obtained by randomly choosing between comparing two model-predicted layouts or comparing a model-predicted layout with the ground truth.

\vspace{1mm}
\noindent
\textbf{Evaluation Protocol:} To assess the effectiveness of AAPA, including pre-training and quality-metric-based filtering, we evaluate two key metrics: \textbf{Mean Intersection over Union (mIoU)} and \textbf{$\mathcal{M}_{\text{judge}}$ win rate}. The mIoU is computed across three different inference settings: \textit{All} (predicting the position of all elements), \textit{Single} (predicting the position of a single text element), and \textit{Multiple} (predicting the position of all text elements). This allows us to evaluate the model's ability to accurately predict layout configurations under varying levels of complexity. The $\mathcal{M}_{\text{judge}}$ win rate is calculated in the \textit{All} inference mode and evaluates the aesthetic quality of generated layouts by comparing them against ground truth using $\mathcal{M_{\text{judge}}}$. This dual-metric evaluation ensures a comprehensive assessment of both spatial accuracy and subjective visual appeal.



\begin{table}[t!]
\centering
\resizebox{0.48\textwidth}{!}{%
\begin{tabular}{lcccc}
\toprule
\multirow{2}{*}{\textbf{Method}} & \multicolumn{3}{c}{\textbf{Mean IoU (\%)}} & \textbf{$\mathcal{M}_{\text{judge}}$ } \\ 
                & \textbf{All}      & \textbf{Single}  & \textbf{Multiple}  & \textbf{Win Rate (\%)} \\ \midrule
SmartText+      & -                 & 4.7              & 2.3                & -   \\        
Typography LMM  & -                 & 40.2             & 17.2               & -       \\    
FlexDM          & 12.71             & 35.5             & 10.3               & 0.93     \\      
LACE            & 23.18             & 41.96            & 21.49              & 3.51      \\  
PosterLLaVa     & 25.18	            & 42.74            & 23.58	            & 5.03          \\
LayoutNUWA      & 25.74                & 43.83               & 24.16                  & 5.58           \\ \midrule
AesthetiQ-1B    & 22.85             & 40.83            & 26.55              & 2.43        \\
AesthetiQ-2B    & 28.19             & 45.92            & 30.44              & 6.13        \\
AesthetiQ-4B    & 38.16             & 49.27            & 37.14              & 14.74       \\
AesthetiQ-8B    & \textbf{42.83 }            & \textbf{52.67}            & \textbf{40.64}              & \textbf{17.19}       \\ 
\bottomrule
\end{tabular}}
\caption{Comparison of layout generation methods based on Mean IoU (\%) and Judge Win Rate (\%) on Crello Dataset. AesthetiQ models outperform baselines, achieving higher IoU and Judge Win Rate, with AesthetiQ-8B showing the best overall performance.}
\label{tab:comparison_crello}
\end{table}

\begin{table}[t!]
\centering
\resizebox{0.44\textwidth}{!}{%
\begin{tabular}{lcccc}
\toprule
\textbf{Method} & \textbf{Mean IoU (\%)} & \textbf{$\mathcal{M}_{\text{judge}}$ Win Rate (\%)} \\ 
\midrule
Desigen            & 15.36            &  4.81        \\  
LACE            &  17.88            &      5.27       \\  
PosterLLaVa     & 	   30.19         &   14.73 \\
LayoutNUWA      &     32.16            &  15.28                \\ 
\midrule
AesthetiQ-1B    &    38.47        &    19.29          \\
AesthetiQ-8B    &  \textbf{48.29}            & \textbf{24.48}             \\ 
\bottomrule
\end{tabular}}
\caption{Comparison of \OURNAME\ with baseline methods based on Mean IoU (\%) and Judge Win Rate (\%) on WebUI Dataset, showcasing significantly superior performance of \OURNAME.}
\label{tab:comparison_webui}
\end{table}

\begin{figure*}[t!]
\centering
  \includegraphics[width=0.9\textwidth]{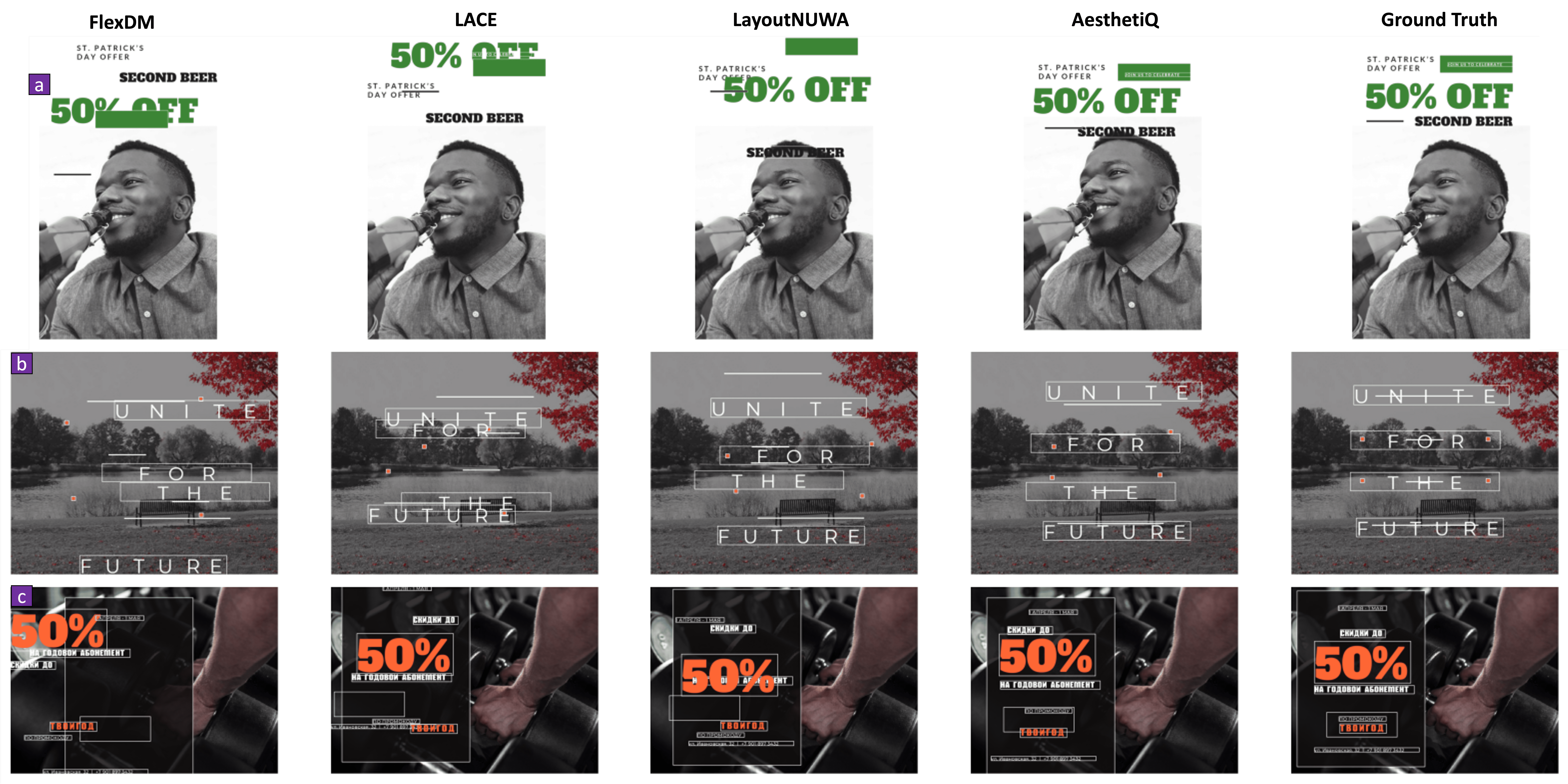}
  \caption{Qualitative comparison of our model, \OURNAME, against recent methods FlexDM, LACE, and LayoutNUWA. Despite the challenge of arranging numerous elements, \OURNAME\ consistently achieves superior layout quality. In row (a), \OURNAME\ effectively places text within salient regions, maintaining clear hierarchy and avoiding overlaps, which enhances readability and aesthetic appeal. In row (b), it achieves precise alignment across elements and optimally positions diverse shapes, preserving a cohesive visual structure. Row (c) showcases \OURNAME's advanced semantic understanding, generating a visually balanced and aesthetically pleasing layout. Overall, \OURNAME\ consistently outperforms competitors in creating coherent, well-structured designs that align with human aesthetic preferences.}
  \label{qualitativeResults}
\end{figure*}

\section{Results and Analysis}
\textbf{Comparison of \OURNAME\ with baselines:} Table \ref{tab:comparison_crello} compares layout generation methods on the \textbf{Crello} dataset, showcasing the superior performance of our \textbf{AesthetiQ} models across multiple metrics. \textbf{AesthetiQ-8B} achieves the highest \textbf{Mean IoU} scores, with 42.83\% in the \textit{All} setting, 52.67\% in the \textit{Single} setting, and 40.64\% in the \textit{Multiple} setting, significantly surpassing baselines like \textbf{LACE} ~\citep{chen2024towards} (23.18\% in \textit{All}) and \textbf{LayoutNUWA} ~\citep{tang2024layoutnuwa} (25.74\% in \textit{All}). Additionally, AesthetiQ-8B achieves the best \textbf{$\mathcal{M}_{\text{judge}}$ win rate} of 17.19\%, highlighting its ability to produce layouts that are both geometrically precise and aesthetically appealing. These consistent improvements in mIoU and judge win rate reflect AesthetiQ's capacity to model intricate design structures while integrating aesthetic constraints, resulting in layouts that better align with judge preferences compared to other methods like \textbf{FlexDM} ~\citep{inoue2023document} and \textbf{SmartText+} ~\citep{9525300}, which perform poorly across both metrics. Moreover, multi-modal LLM-based methods such as PosterLLaVa ~\citep{yang2024posterllava} and LayoutNUWA also fall short of AesthetiQ-8B, underscoring the robustness of our approach in handling diverse layout configurations while maintaining high visual quality.

Table \ref{tab:comparison_webui} further demonstrates that \OURNAME\ models significantly outperform all baselines on the WebUI dataset across both metrics. For instance, AesthetiQ-8B achieves a remarkable Mean IoU of 48.29\% and a judge win rate of 24.48\%, well above LayoutNUWA, the top-performing baseline, which scores 32.16\% in Mean IoU and 15.28\% in win rate. Even smaller variants, like AesthetiQ-1B, exhibit strong performance with a Mean IoU of 38.47\% and a judge win rate of 19.29\%. In contrast, previous methods such as Desigen~\citep{10655522} (15.36\% IoU, 4.81\% win rate) and LACE (17.88\% IoU, 5.27\% win rate) perform significantly worse. These results highlight AesthetiQ's strength in producing layouts that are not only more accurate but also more aesthetically pleasing.

\noindent
\textbf{Qualitative Results:} Figure \ref{qualitativeResults} presents a qualitative comparison of \OURNAME\ against strong baselines. In row (a), \OURNAME\ skillfully places text in prominent areas, establishing a clear hierarchy and avoiding overlaps, which improves readability and visual appeal. Row (b) highlights \OURNAME's precise alignment of elements and optimal positioning of various shapes, resulting in a unified visual layout. In row (c), \OURNAME\ demonstrates its advanced semantic understanding, producing a balanced and aesthetically pleasing design. Overall, \OURNAME\ consistently surpasses baselines by generating coherent, well-structured layouts that align with human aesthetic preferences. A user study, detailed in the Appendix, reveals that \OURNAME\ was consistently the top choice for predicted layouts when compared to FlexDM, LACE, LayoutNUWA, and human preferences correlate well with the preference distribution of judge model $M_{\text{judge}}$.

\begin{figure*}[t]
  \includegraphics[width=0.95\textwidth]{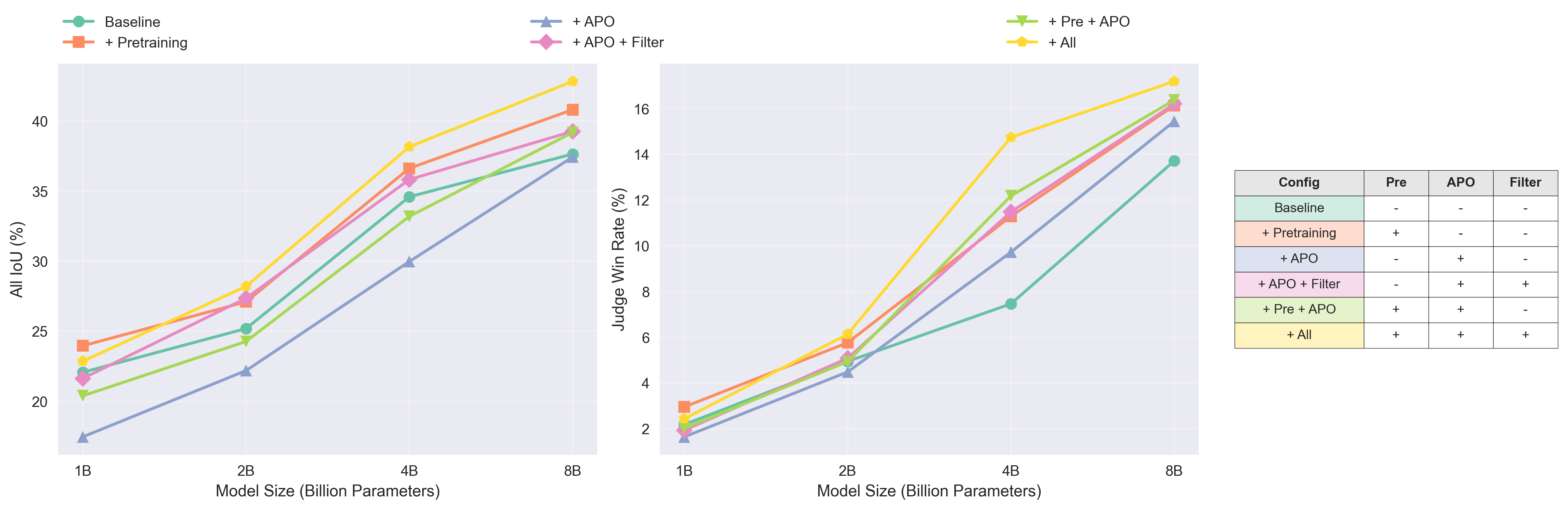}
  \caption{Performance improvement across scale (1B–8B parameters) for layout generation, showing effects of pretraining, quality filtering, and Aesthetic-Aware Preference Alignment (AAPA). \textbf{Left:} IoU progression under different training configurations. \textbf{Middle:} $\mathcal{M_{\text{judge}}}$ Win Rate improvements, emphasizing the impact of AAPA and pretraining. \textbf{Right:} Configuration table indicating settings for each experiment. The results underscore the impact of each design component in \OURNAME, emphasizing their role in tackling layout generation challenges.}
  \label{analysisDiagram}
\end{figure*}

\noindent
\textbf{Larger Models Capture Aesthetics Better:} Scaling up model size significantly improves layout generation performance. As seen in Fig \ref{analysisDiagram}, larger models outperform smaller ones in the metrics \textit{All}, \textit{Single}, \textit{Multiple} IoU, and Judge Win Rate. For example, moving from 1B to 8B parameters boosts the Judge Win Rate from 2.43\% to 17.19\%, suggesting larger models better capture aesthetic nuances and complex spatial relationships. This trend indicates a strong alignment between model scale and the ability to reflect aesthetic preferences, making larger models more suited for aesthetic-aware layout generation. Further, this also showcases the difficulty of the layout generation task.

\noindent
\textbf{Pre-training Enhances Layout Understanding:} Pretraining with a large dataset of 80,000 templates enhances the model's grasp of layout structuring, providing a strong foundation for fine-tuning and AAPA. Models pretrained on this dataset show improvements across metrics. For instance, pretraining raises the 8B model’s IoU from 37.64\% to 40.81\%. This phase equips models with essential layout generation and understanding capabilities, which AAPA further refines to achieve aesthetic alignment.

\noindent
\textbf{Quality-metrics Based Layout Filtering Improves AAPA Training:} Quality filtering, guided by alignment maximization and overlap minimization heuristics, enhances model performance by selecting high-quality layouts for training. Without filtering, aesthetic quality and layout coherence decline, as suboptimal samples impact learning. AAPA’s performance drops without filtering, with decreases in IoU across scales. Quality filtering boosts the 4B model’s IoU by 4.97\% and Judge Win Rate by 2.56\%, highlighting its importance in maintaining aesthetic standards.

\vspace{1.5mm}
\noindent
\textbf{Effect of AAPA on Performance:} AAPA considerably enhances the aesthetic quality of generated layouts by aligning them with preferences derived from Multi-modal Large Language Models (VLMs). This preference-based alignment improves the Judge Win Rate, guiding the model towards aesthetically pleasing layouts rather than exact ground truth matches. \OURNAME-1B model shows limited gains, as it faces challenges in producing high-quality layouts during sampling, limiting the effectiveness of preference optimization. However, AAPA-trained larger models demonstrate superior performance on Crello and WebUI benchmarks across metrics, underscoring the value of preference-based training in aesthetic-sensitive tasks.

\begin{figure}[t]
    \centering
    \includegraphics[width=0.8\linewidth]{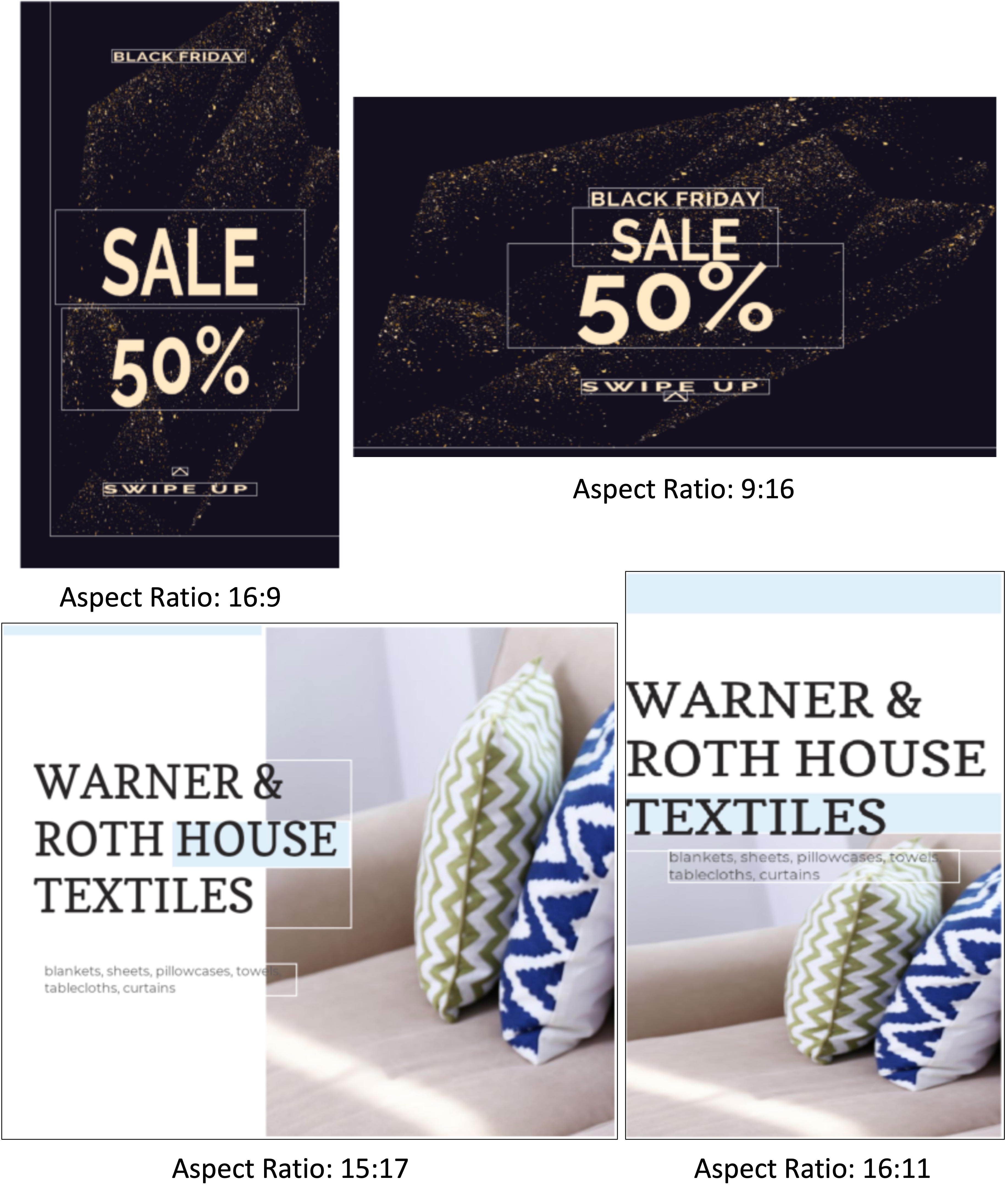}
    \caption{Capability of \OURNAME\ to generate templates in various aspect ratios by changing the canvas height and width}
    \label{aspectFig}
\end{figure}

\vspace{1.5mm}
\noindent
\textbf{Generating Layouts with Different Aspect Ratios:} Layouts must adapt to various aspect ratios across platforms. Our model maintains stable performance across common aspect ratios like 4:3, 16:9, and square formats, effectively rearranging elements to fit different dimensions. As shown in Fig \ref{aspectFig}, the model consistently produces visually balanced layouts for diverse aspect ratios, demonstrating its flexibility and robustness in generating aesthetically coherent layouts across various screen requirements.

\section{Conclusion} In this work, we introduced \textbf{Aesthetic-Aware Preference Alignment} (AAPA), a novel approach for training Multi-modal Large Language Models (MLLMs) in the task of aesthetic-driven layout prediction. By leveraging aesthetic preferences directly through a preference optimization mechanism, AAPA addresses critical limitations of existing generative models that lack contextual aesthetic understanding and alignment with human-like preferences. Our proposed data filtering protocol, based on layout-quality heuristics, ensures that only high-quality layouts contribute to training, further enhancing model performance. Additionally, we introduced a novel evaluation metric to assess the aesthetic superiority of generated layouts against ground-truth designs, supporting a more holistic evaluation of layout quality.

{
    \small
    \bibliographystyle{ieeenat_fullname}
    \bibliography{main}
}
 

\setcounter{page}{1}
\maketitlesupplementary
\begin{figure*}[t!] 
    \centering
    \includegraphics[width=\textwidth]{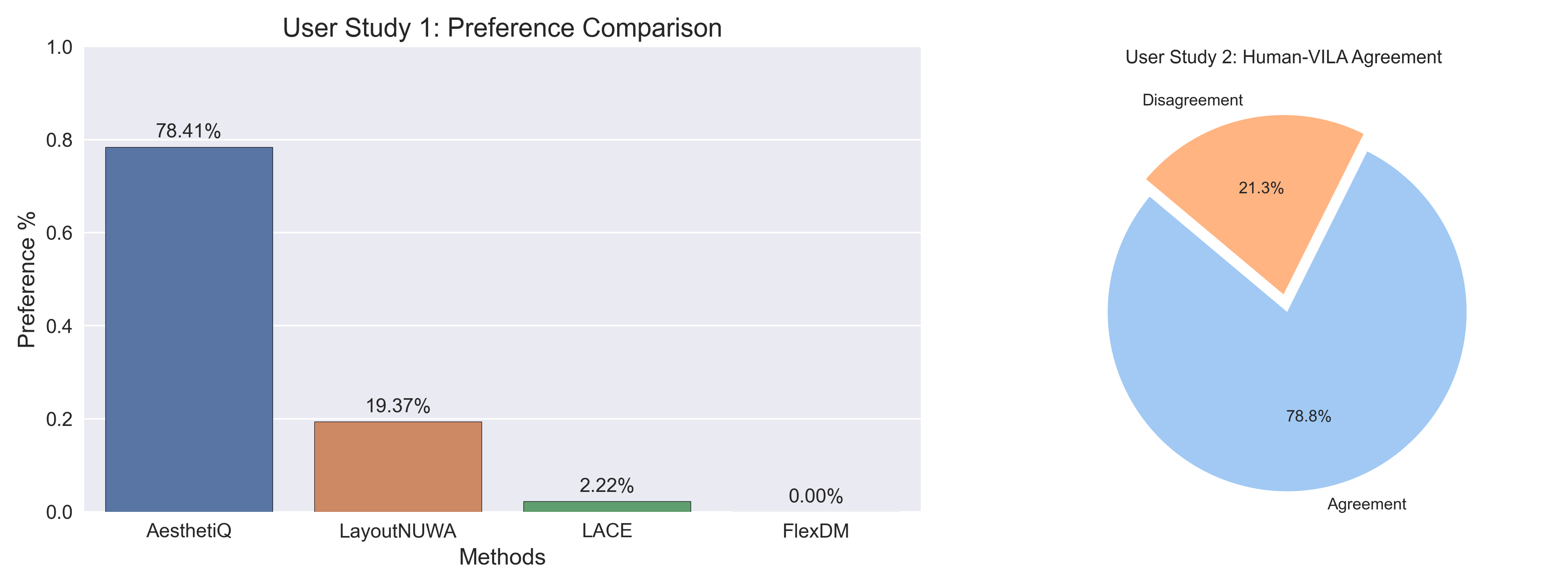} 
    \caption{Comparison of results from two user studies evaluating layout aesthetics. The bar plot (left) shows the preference of layout predictions across four methods: AesthetiQ, LayoutNUWA, LACE, and FlexDM, with AesthetiQ significantly outperforming others. The pie chart (right) evaluates alignment between human preferences and the VILA model, achieving a substantial agreement rate of 78.8\%. Together, these results highlight the superiority of AesthetiQ in generating aesthetically pleasing layouts and the reliability of VILA as an evaluator.}
    \label{fig:wide}
\end{figure*}

\maketitle


\section{User Study}


To evaluate the aesthetic quality of layouts generated by various methods and validate the alignment of human preferences with our approach, we conducted two user studies involving 22 diverse volunteers. The participants were selected to represent a broad spectrum of demographics, including variations in age, gender, occupation, and religion, ensuring a well-rounded and inclusive evaluation. Each participant was presented with a total of 30 questions, designed to capture their aesthetic preferences and opinions on the generated layouts.

\paragraph{User Study 1:} Participants were shown layout predictions from four methods: AesthetiQ, LayoutNUWA, LACE, and FlexDM. They were instructed to select the better layout based on aesthetics, alignment and overlap between text and images, and whether the text in the layout made sense. The results are as follows:
\begin{itemize}
    \item \textbf{AesthetiQ}: 78.41\%
    \item \textbf{LayoutNUWA}~\citep{tang2024layoutnuwa}: 19.37\%
    \item \textbf{LACE}~\citep{chen2024towards}: 2.22\%
    \item \textbf{FlexDM}~\citep{inoue2023document}: 0.00\%
\end{itemize}

These results highlight the significant preference for layouts generated by AesthetiQ compared to the baselines, underscoring its ability to produce more visually appealing and coherent designs.

\paragraph{User Study 2:} Participants were shown pairs of model-generated layouts and asked to select the better one using the same instructions as in the first study. The same layouts were also evaluated by VILA, a Vision-Language Model (VLM) judge. We measured the agreement between human preferences and VILA's outputs, which yielded an alignment score of 78.8\%.

This result demonstrates that VILA’s aesthetic judgment aligns well with human preferences, further validating its use as an aesthetic evaluator in our framework.

\section{Complete Results on WebUI}
\begin{table}[t!]
\centering
\resizebox{0.44\textwidth}{!}{%
\begin{tabular}{lcccc}
\toprule
\textbf{Method} & \textbf{Mean IoU (\%)} & \textbf{$\mathcal{M}_{\text{judge}}$ Win Rate (\%)} \\ 
\midrule
Desigen~\citep{10655522}            & 15.36            &  4.81        \\  
LACE~\citep{chen2024towards}            &  17.88            &      5.27       \\  
PosterLLaVa~\citep{yang2024posterllava}     & 	   30.19         &   14.73 \\
LayoutNUWA~\citep{tang2024layoutnuwa}     &     32.16            &  15.28                \\ 
\midrule
AesthetiQ-1B    &    38.47        &    19.29          \\
AesthetiQ-2B    &   41.42           &  21.87          \\
AesthetiQ-4B    &  44.16           &  22.74            \\
AesthetiQ-8B    &  \textbf{48.29}            & \textbf{24.48}             \\ 
\bottomrule
\end{tabular}}
 \caption{Comparison of \OURNAME\ with baseline methods on the WebUI dataset, evaluated using Mean IoU (\%) and $\mathcal{M}{\text{judge}}$ Win Rate (\%). The results demonstrate the superior performance of \OURNAME\ across all model scales, with notable gains in aesthetic and structural alignment metrics.}
\label{tab:comparison_webui}
\end{table}

Table \ref{tab:comparison_webui} provides a detailed comparison of our approach, \OURNAME, against baseline methods, including Desigen, LACE, PosterLLaVa, and LayoutNUWA. While previous methods like PosterLLaVa and LayoutNUWA achieve decent performance, they fall short in terms of both structural coherence and aesthetic alignment. In contrast, \OURNAME\ shows consistent improvements across all metrics, achieving the highest Mean IoU and $\mathcal{M}_{\text{judge}}$ Win Rate.

We observe a clear trend of performance scaling with model size. The Mean IoU improves progressively from 38.47\% for the 1B model to 48.29\% for the 8B model.
Similarly, the Judge Win Rate increases from 19.29\% to 24.48\%, showcasing the model's alignment with human aesthetic preferences as the scale grows. In the main paper, due to space constraints, we presented results for only the 1B and 8B variants of \OURNAME. Here, we include results for the 2B and 4B variants to offer a comprehensive analysis of our model's performance across different scales. The full results underscore the scalability and effectiveness of our approach, particularly in leveraging aesthetic preferences to optimize layout quality.These findings highlight the robustness of \OURNAME\ in addressing the challenges of layout generation, establishing a new benchmark for performance on the WebUI dataset. 

In the paper, we primarily focus on showcasing qualitative results on the Crello dataset, as it contains individual elements, allowing for detailed analysis and visualization. In contrast, the WebUI dataset only includes category labels and their positions, making it impossible to generate the final rendered templates. For the AAPA evaluation, we render the bounding boxes of the predicted elements on a background, similar to the approach used in Desigen~\citep{10655522}. These renderings are then evaluated by the judge VLM, which selects the layout it deems superior between the two.


\begin{table}[h!]
\centering
\resizebox{0.5\textwidth}{!}{%
\begin{tabular}{lcccccc}
\toprule
\textbf{AAPA $\mathcal{A}_{\text{judge}}$} & \multicolumn{3}{c}{\textbf{Mean IoU (\%) $\uparrow$}} & \textbf{Eval $\mathcal{M}_{\text{judge}}$} & \textbf{$\mathcal{M}_{\text{judge}}$ Win Rate (\%) $\uparrow$} \\ 
\cmidrule(lr){2-4}
 & \textbf{All} & \textbf{Single} & \textbf{Multiple} &  &  \\ 
\midrule
\multirow{2}{*}{VILA (Paper)}            & \multirow{2}{*}{42.83} & \multirow{2}{*}{52.67} & \multirow{2}{*}{40.64} & Vila (Paper) &  17.19        \\  
            &                        &                        &                        & Gpt4o  &   14.27           \\  
\midrule
\multirow{2}{*}{Gpt4o}           & \multirow{2}{*}{\textbf{44.79}} & \multirow{2}{*}{\textbf{55.81}} & \multirow{2}{*}{\textbf{43.28}} & Vila   &   \textbf{19.41}           \\
           &                        &                        &                        & Gpt4o  & \textbf{15.74} \\ 
\bottomrule
\end{tabular}}
\caption{Comparison of AesthetiQ-8B with different training ($\mathcal{A}_{\text{judge}}$) \& eval ($\mathcal{M}_{\text{judge}}$) judges. mIoU is independent of $\mathcal{M}_{\text{judge}}$.}
\label{tab:diffJudge}
\end{table}


\section{Stronger MLLM training} 

We chose VILA-7B as the judge for its open-source, license-friendly nature. Training and evaluation with GPT-4o (Tab \ref{tab:diffJudge}) improved all metrics with consistent trends in $\mathcal{M}_{\text{judge}}$ win rate across ablations (Paper Fig. 4).


\paragraph{Justification for MLLM as judge:} We conduct a user study to measure VILA's correlation with human aesthetic preferences, finding 78.8\% agreement. GPT-4o achieves \textbf{88.6\%} correlation, \& AesthetiQ-8B performs better with a stronger judge (Tab \ref{tab:diffJudge}).

\section{Detailed Experimental Results}
\label{appendix:results}

This section provides the complete experimental results referenced in the main paper, presented in Table \ref{tab:internvl_performance_no_judge_epochs}. The table details the performance of our models across various configurations, highlighting the effects of scaling, pretraining, VILA alignment, and quality filtering on layout generation tasks. Metrics include All IoU, Single Text IoU, Multiple Text IoU, and Judge Win Rate. These results support the analysis presented in Section 5 of the main paper, showcasing the effectiveness of Aesthetic-Aware Preference Alignment (AAPA) and other components in enhancing the quality and alignment of generated layouts.

\begin{table*}[ht]
\centering
\resizebox{\textwidth}{!}{%
\begin{tabular}{lllllcccc}
\toprule
\multirow{2}{*}{\textbf{Method}} & \multirow{2}{*}{\textbf{LLM}} & \multirow{2}{*}{\textbf{Pretraining}} & \multirow{2}{*}{\textbf{VILA Alignment}} & \multirow{2}{*}{\textbf{Data Filtering}} & \multicolumn{3}{c}{\textbf{Mean IoU}} & \multirow{2}{*}{$\mathcal{M}_{\text{judge}}$\textbf{Win Rate (\%)}} \\
& & & & & \textbf{All} & \textbf{Single} & \textbf{Multiple} & \\
\midrule
\OURNAME\ -1B     & Qwen-0.5b    & No                   & No                     & No                     & 22.06           & 40.14                   & 24.88                     & 2.18                         \\
\OURNAME\ -1B     & Qwen-0.5b    & Yes                  & No                     & No                     & 23.95           & 42.19                   & 26.93                     & 2.95                         \\
\OURNAME\ -1B     & Qwen-0.5b    & No                   & Yes                    & No                     & 17.45           & 35.91                   & 20.76                     & 1.64                         \\
\OURNAME\ -1B     & Qwen-0.5b    & No                   & Yes                    & Yes                    & 21.62           & 39.56                   & 25.03                     & 1.93                         \\
\OURNAME\ -1B     & Qwen-0.5b    & Yes                  & Yes                    & No                     & 20.38           & 38.24                   & 23.92                     & 2.02                         \\
\OURNAME\ -1B     & Qwen-0.5b    & Yes                  & Yes                    & Yes                    & 22.85           & 40.83                   & 26.55                     & 2.43                         \\
\midrule
\OURNAME\ -2B     & InternLM-1.8b & No                  & No                     & No                     & 25.18           & 43.28                   & 26.94                     & 4.94                         \\
\OURNAME\ -2B     & InternLM-1.8b & Yes                 & No                     & No                     & 27.09           & 44.61                   & 28.94                     & 5.76                         \\
\OURNAME\ -2B     & InternLM-1.8b & No                  & Yes                    & No                     & 22.18           & 41.64                   & 24.14                     & 4.48                         \\
\OURNAME\ -2B     & InternLM-1.8b & No                  & Yes                    & Yes                    & 27.35           & 44.81                   & 29.44                     & 5.08                         \\
\OURNAME\ -2B     & InternLM-1.8b & Yes                 & Yes                    & No                     & 24.26           & 43.83                   & 26.44                     & 4.93                         \\
\OURNAME\ -2B     & InternLM-1.8b & Yes                 & Yes                    & Yes                    & 28.19           & 45.92                   & 30.44                     & 6.13                         \\
\midrule
\OURNAME\ -4B     & Phi3-3.8b     & No                  & No                     & No                     & 34.59           & 47.61                   & 33.19                     & 7.46                         \\
\OURNAME\ -4B     & Phi3-3.8b     & Yes                 & No                     & No                     & 36.62           & 48.32                   & 35.47                     & 11.29                        \\
\OURNAME\ -4B     & Phi3-3.8b     & No                  & Yes                    & No                     & 29.97           & 44.48                   & 31.14                     & 9.72                         \\
\OURNAME\ -4B     & Phi3-3.8b     & No                  & Yes                    & Yes                    & 35.82           & 47.65                   & 34.93                     & 11.48                        \\
\OURNAME\ -4B     & Phi3-3.8b     & Yes                 & Yes                    & No                     & 33.19           & 46.94                   & 33.42                     & 12.18                        \\
\OURNAME\ -4B     & Phi3-3.8b     & Yes                 & Yes                    & Yes                    & 38.16           & 49.27                   & 37.14                     & 14.74                        \\
\midrule
\OURNAME\ -8B     & InternLM-7b   & No                  & No                     & No                     & 37.64           & 51.01                   & 36.32                     & 13.71                        \\
\OURNAME\ -8B     & InternLM-7b   & Yes                 & No                     & No                     & 40.81           & 51.82                   & 38.51                     & 16.13                        \\
\OURNAME\ -8B     & InternLM-7b   & No                  & Yes                    & No                     & 37.43           & 48.48                   & 34.18                     & 15.44                        \\
\OURNAME\ -8B     & InternLM-7b   & No                  & Yes                    & Yes                    & 39.26           & 51.15                   & 38.11                     & 16.20                        \\
\OURNAME\ -8B     & InternLM-7b   & Yes                 & Yes                    & No                     & 39.18           & 50.34                   & 36.42                     & 16.37                        \\
\OURNAME\ -8B     & InternLM-7b   & Yes                 & Yes                    & Yes                    & 42.83           & 52.67                   & 40.64                     & 17.19                        \\
\bottomrule
\end{tabular}%
}
\caption{Performance of \OURNAME\ across scales (1B, 2B, 4B, 8B) on the Crello dataset, evaluating the effects of pretraining, VILA alignment, and data filtering on IoU metrics and judge win rates. The results demonstrate the scalability and effectiveness of the aesthetic-aware preference alignment method.}
\label{tab:internvl_performance_no_judge_epochs}
\end{table*}

\section{Direct Preference Optimization}
Direct Preference Optimisation (DPO)~\citep{NEURIPS2023_a85b405e} emerged as an alternative approach to Reinforcement Learning using Human Feedback (RLHF)~\citep{NEURIPS2022_b1efde53}, eliminating the requirement of training a reward model. While RLHF relies on a reward model to evaluate LLM outputs for fine-tuning through reinforcement learning to achieve human preference alignment, DPO takes a different approach. It converts the reward-function loss into a loss over the LLM policy, enabling implicit reward optimization through policy loss optimization. This is achieved using human preference data that pairs two LLM-generated outputs, where one is designated as the winner candidate - $y_w$ and the other as the loser candidate - $y_l$. Using a static dataset structured as $\mathcal{D} = \{x, y_w, y_l\}$, where x represents the input, the loss is formulated as:

\begin{align}
    \mathcal{L}_{R} = -log[\sigma(r(x, y_w) - r(x, y_l))] \\
    r(x, y) = \beta log(\frac{\pi_{\theta}(y|x)}{\pi_{ref}(y|x)})
\end{align}

Here, $\pi_{\mathcal{Z}}(y | x)$ denotes the probability of generating $y$ given input $x$ for model $\mathcal{Z} \in \{\mathcal{M}_{ref}, \mathcal{M}_{\theta}\}$, where $\mathcal{M}_{ref}$ typically represents the instruction fine-tuned model for LLMs to maintain policy proximity to the initial model, and $\mathcal{M}_{\theta}$ represents the LLM policy being optimized through DPO. Additionally, $\sigma$ represents the sigmoid activation, and $\beta$ is a parameter controlling the deviation extent from the reference model. In essence, this algorithm trains the LLM to develop output preferences among candidates without explicitly modeling rewards. Our algorithm \textit{Aesthetic-Aware Preference Alignment} (AAPA) draws motivation from DPO and carries out preferential training across different layout configurations. For a more comprehensive understanding on DPO, readers are directed to the original publication~\citep{NEURIPS2023_a85b405e}.

\section{Prompt Templates for Layout and Judge VLMs}
The following prompt template was used as input to our layout generation model $\mathcal{M}_{\text{layout}}$ to guide the generation of aesthetic poster layouts. The template specifies the canvas dimensions and provides a structured description of the elements to be placed, including their type (e.g., text or image), content, and category. This format allows the model to interpret the spatial constraints and semantic attributes of each element effectively, enabling systematic exploration of layout generation. The \texttt{\textless image\textgreater} token in the prompt is replaced with a sequence of image embeddings corresponding to the input images, ensuring that the model processes visual information in a compact and meaningful way. By explicitly defining these attributes, the template facilitates reproducibility and evaluation of layout designs. The prompt template is shown below:

\vspace{0.5cm}
\begin{ttfamily}
\noindent Consider the image \textless image\textgreater\ with height and width of \{canvas\_height\} and \{canvas\_width\}. The following elements need to be placed on the image to obtain an aesthetic poster layout.

\medskip

Element 1: \\
Text: LOREM IPSUM \\
Category: text

\medskip

Element 2: \\
Image: \textless image\textgreater \\
Category: image

...
\end{ttfamily}

This structured input format ensures that the model can accurately process both visual and textual elements while adhering to aesthetic principles, making it particularly suitable for tasks in computer vision and graphics. 

The following prompt is used as input to the judge visual language model \(\mathcal{M_{\text{judge}}}\) to evaluate and compare two visual templates based on predefined criteria: aesthetics, clarity, usability, creativity, and consistency. The model processes these criteria to determine which template is superior and outputs the result in a structured JSON format: \texttt{\{"better\_layout": "answer"\}}, where the answer specifies the preferred template (\texttt{image\_1} or \texttt{image\_2}). This structured approach ensures objective and standardized evaluation of visual designs. The prompt is shown below:

\medskip
\begin{ttfamily}
You are a visual language model designed to evaluate and rate visual templates. You are presented with 2 visual templates, and your task is to choose the better template between these 2 based on the following criteria: 

\medskip
\noindent
Aesthetics: How visually appealing is the template, \\
\noindent
Clarity: How clear and easy to understand is the template, \\
\noindent
Usability: How practical and user-friendly is the template, \\
\noindent
Creativity: How unique and innovative is the design, \\
\noindent
Consistency: How consistent is the template with design principles and standards. 

\medskip
\noindent
Please provide your answer in the following JSON format and do not include any other details: 

\medskip
\noindent
\{"better\_layout": "answer"\} 

\medskip
\noindent
where answer could either be \texttt{image\_1} or \texttt{image\_2}.
\end{ttfamily}

\section{Qualitative results}

Due to limited space, we included only a few examples of comparison in the main paper. In the following pages, we show more examples for a more comprehensive comparison.

\begin{figure*}[t]
  \includegraphics[width=\textwidth]{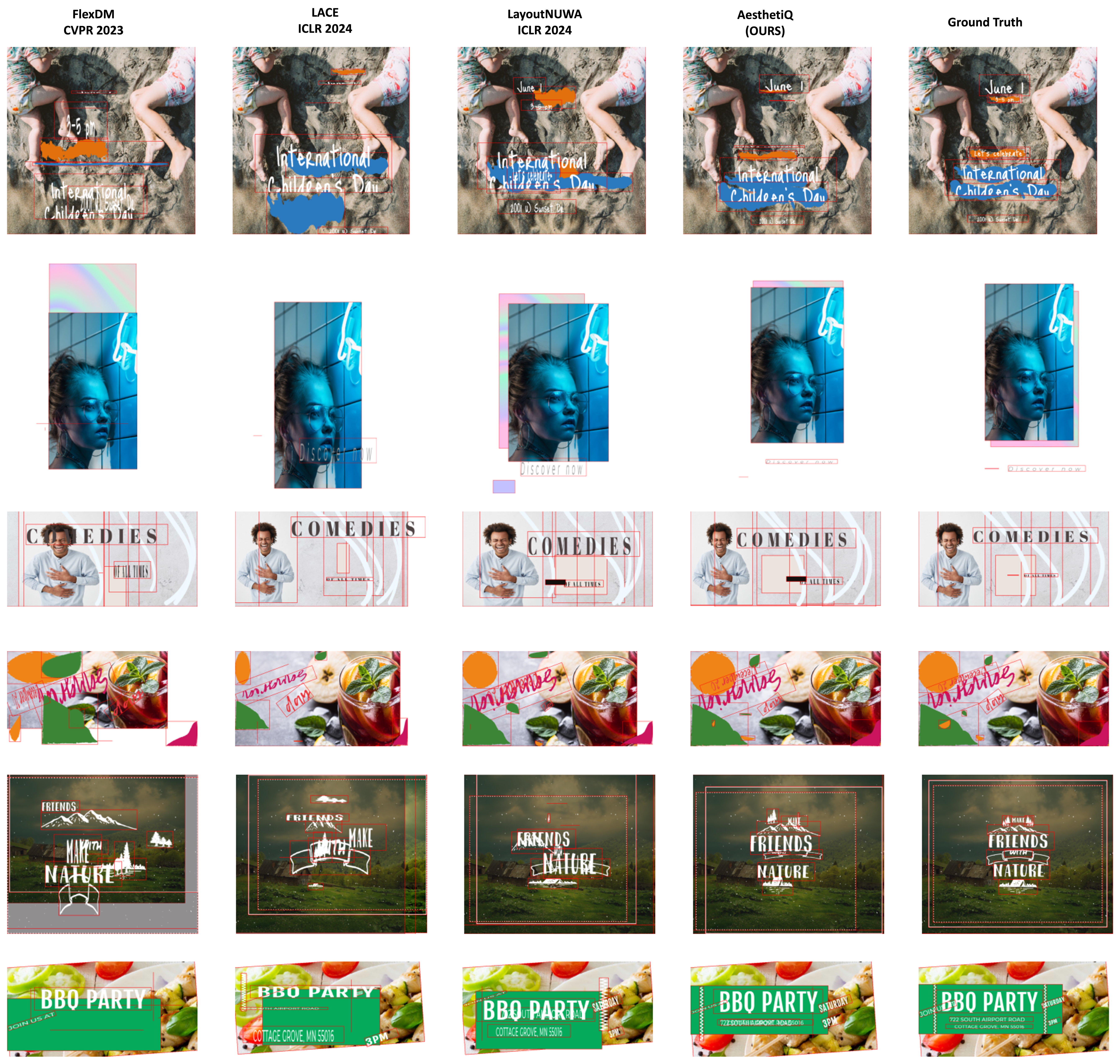}
  \caption{Qualitative comparison of various baselines for layout prediction}
\end{figure*}



\end{document}